\documentclass{article}

\usepackage{arxiv}

\usepackage[utf8]{inputenc} 
\usepackage[T1]{fontenc}    
\usepackage{hyperref}       
\usepackage{url}            
\usepackage{booktabs}       
\usepackage{amsfonts}       
\usepackage{nicefrac}       
\usepackage{microtype}      
\usepackage{lipsum}
\usepackage{graphicx}
\usepackage{epsfig}
\usepackage{subcaption}
\usepackage{amsmath,amssymb} 
\usepackage[ruled, vlined]{algorithm2e}
\usepackage{algorithmic}

\title{AutoPruning for Deep Neural Network with Dynamic Channel Masking}

\author{Baopu Li\textsuperscript{1*}, Yanwen Fan\textsuperscript{2*}, Zhihong Pan\textsuperscript{1}, Gang Zhang \textsuperscript{2} \\
*means equal contribution,\\
\textsuperscript{1}Baidu Research, Sunnyvale, CA \\
          \textsuperscript{2}Department of Vision Technology, Baidu Incorporation
}

\begin{document}
\newcommand{\etal}{\textit{et al}.}

\maketitle
\vspace{18pt}
\begin{abstract}
Modern deep neural network models are large and computationally intensive. One typical solution to this issue is model pruning. However, most current pruning algorithms depend on hand crafted rules or domain expertise. To overcome this problem,  we propose a learning based auto pruning algorithm for deep neural network, which is inspired by recent automatic machine learning(AutoML).  
A two objectives' problem that aims for the the weights and the best channels for each layer is first formulated. 
An alternative optimization approach is then proposed to derive the optimal channel numbers and weights simultaneously. In the process of pruning,  we utilize a searchable hyperparameter, remaining ratio, to denote the number of channels in each convolution layer, and then a dynamic masking process is proposed to describe the corresponding channel evolution.
To control the trade-off between the accuracy of a model and the pruning ratio of floating point operations, a novel loss function is further introduced. Preliminary experimental results on benchmark datasets demonstrate that our scheme achieves competitive results  for neural network pruning.  
\end{abstract}


\section{Introduction}

 Deep neural network(DNN) models have greatly improved the state-of-the-art for a lot of computer vision tasks such as image classification, image segmentation, objection detection, human pose detection and so on. However, for resource limited mobile and edge devices that are almost ubiquitous nowadays, the large size and computational burden of DNN models seem to be a big challenge to overcome. For instance, the popular VGG-16 model~\cite{2014arXiv1409.1556S} is about 150Mb and needs about 15G floating point operations (FLOPs) to classify one color image of 224* 224~\cite{DBLP:journals/corr/abs-1810-00736}. To alleviate this problem, a lot of efforts along the direction of model compression have been paid and achieved impressive results in recent years. 

A common approach among these wonderful works is to prune the over parameterized DNN models. In the early stage of network model pruning, many researchers devoted themselves to weight pruning or neuron pruning~\cite{NIPS1989_250,DBLP:journals/corr/GuoYC16,DBLP:journals/corr/AghasiNR16} with the aim of sparse connection in the whole network, yielding a good compression ratio. However, the sparse network structure will not necessarily lead to efficiency because it may not be friendly to the hardware. As such, a new pruning trend that directly applies to the filters, also known as structural pruning, is witnessed recently. Wonderful pruning performance together with fast running speed are reported \cite{DBLP:journals/corr/LuoWL17,DBLP:journals/corr/abs-1708-06519,DBLP:journals/corr/LiKDSG16}.

\begin{figure}[t]
\centering
\includegraphics[height=7.12cm]{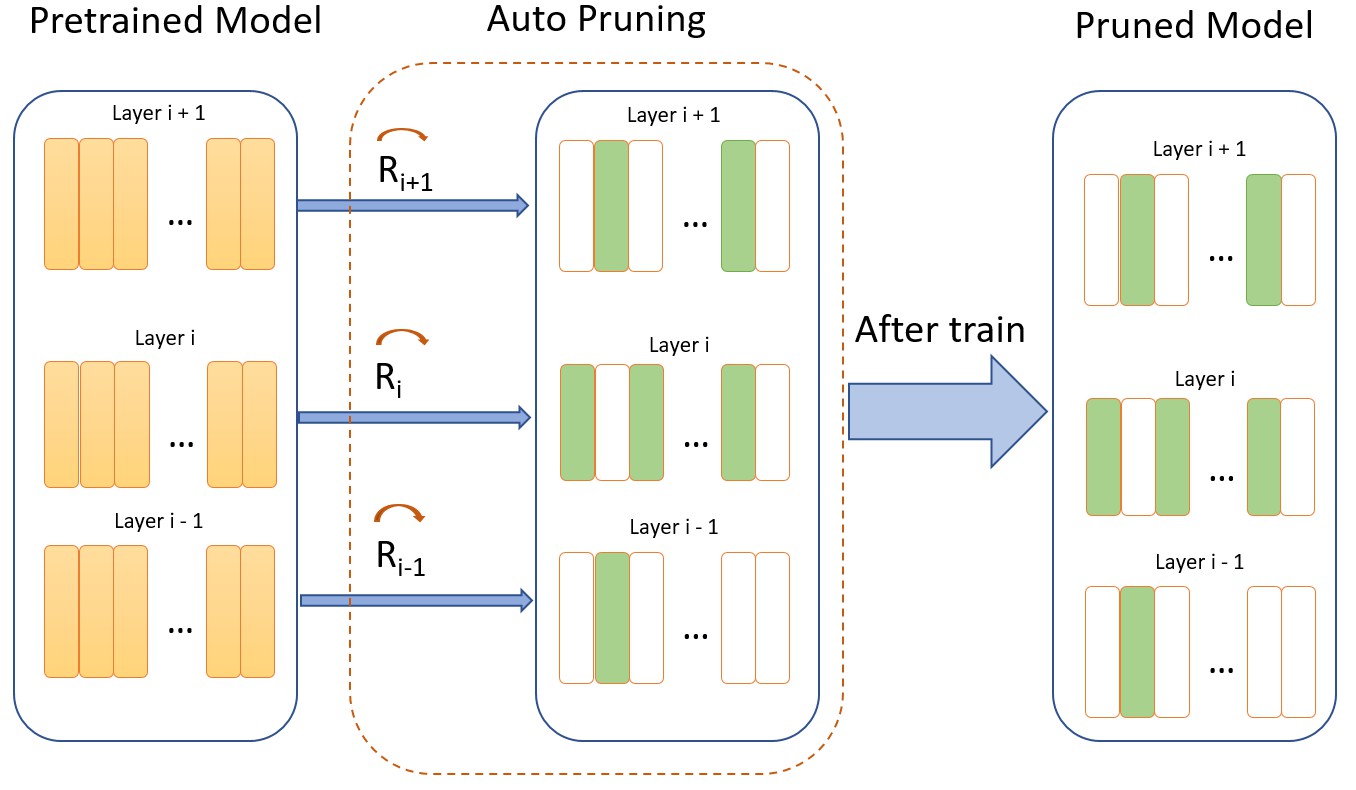}
\caption{Overview of the proposed auto-pruning process for DNN. A channel remaining ratio $R_{i}$ is utilized to denote the channels after pruning in each iteration of search, and it will be updated together with the weight training process of the network model. The curved arrow on each $R_{i}$ means that it is iterable. A dynamic masking process based on $R_{i}$ is also included to output the concrete channel index in the pruning process. The light green color means that this channel is kept in the iterative pruning process. The whole process is described with a two objectives' optimization framework and the final pruned model will be fine tuned after finishing the iterative pruning process.}
\label{fig:pipeline}
\end{figure}

Most of the previous works in the field of pruning for DNN depend on some domain expertise.  In addition, most of them apply hard filter pruning that removes the pruned filters. However, the pruned filters may still contain some useful features for object recognition or classification.  To mitigate these problems, we try to understand the network model pruning from the standpoint of AutoML, designing an efficient learning based channel pruning algorithm. We attempt to automatically search the desired channel number for each layer. To achieve this goal, we first formulate the DNN pruning as a two objectives' problem of simultaneously searching for the optimal weight and hyperparameters of remaining ratio of channels for each layer. Then an alternative gradient descent optimization procedure is leveraged to obtain the solutions. While in this iteration process, we apply a remaining ratio to represent the number of channels in each convolution layer, and then a dynamic masking step based on the importance of the channels is designed to represent the channel evolution process. To control the pruning procedure more effectively, a new loss function that is based on the FLOPs and prediction accuracy is further designed. The outline of our algorithm is described in Fig.\ref{fig:pipeline}. We test our novel algorithm on some benchmarks and show its promising performances.

To summarize, the contributions of our work are as follows:

\begin{itemize}
\item Model pruning is approached from the standpoint of AutoML and we propose a novel two objectives' framework to train the remaining ratio of channels along with the network weights.
\item A novel pruning rule that is based on dynamic channel masking is embedded effectively in the optimization process, leading to efficient channel pruning together with the model training.
\item We suggest a new FLOPs based cost function that controls the trade-off between model accuracy and pruning ratio.
\item Comprehensive experiments demonstrate the effectiveness of the proposed scheme.
\end{itemize}

\section{Related Works}

Model compression has drawn a lot of attention since the popularity of DNN. Roughly speaking, these works can be categorized into four classes:  model pruning, low-rank decomposition, compact convolutional filters and knowledge distillation~\cite{DBLP:journals/corr/abs-1710-09282}. We mainly concentrate on model pruning related efforts in this paper.

\subsection{Model pruning}

Han \etal~\cite{2015arXiv151000149H} came up with an effective network pruning method by thresholding the weight of the neuron connections, and the accuracy of the network is preserved with a promising pruning ratio. They continued to leverage reinforcement learning that is based on Deep Deterministic Policy Gradients(DDPG) agent to automatically prune a DNN and achieved new state-of-the-art of pruning on VGG, ResNet and MobileNet~\cite{DBLP:journals/corr/abs-1802-03494}. 
Liu \etal~\cite{DBLP:journals/corr/abs-1810-05270} advocate that it is the whole pruned network structures rather than the weights play a more important role in the model compression process and triggered deep thoughts about the whole work in this field. Inter-layer dependency is utilized in~\cite{DBLP:journals/corr/abs-1812-00353} and a novel layer-wise recursive Bayesian pruning approach is introduced together with significant acceleration.
In~\cite{IEEE:CVPR/2019/Molchanov19}, the authors proposed an importance estimation scheme that is based on the loss change to describe a channel's contribution for deep neural network and reported impressive performance on ResNet. An interesting dynamic network pruning method~\cite{Liu2018FrequencyDomainDP} is put forward in the frequency domain to exploit the spatial correlations. To achieve a good global compression ratio, Ding \etal ~\cite{2019arXiv190912778D}proposed a new momentum-SGC-based optimization approach with online pruning. The authors designed a set of trainable auxiliary parameter to avoid the instability and noise problem in typical network pruning~\cite{NIPS2019_9521}.
Norm based importance criterion may lose its effects, so the authors in ~\cite{DBLP:journals/corr/abs-1811-00250} proposed a novel channel pruning algorithm  based on geometric median aiming for redundancy reduction.

In contrast to all the above pruning works, our scheme applies learning based mechanism to automatically obtain the optimal channels, thus yielding better performance of pruning without domain expertise. In addition, the pruned filters may still participate the future channel pruning process, effectively keeping a model's capacity while pruning. Compared to the automatic pruning algorithm that is based on reinforcement learning (RL) method in \cite{DBLP:journals/corr/abs-1802-03494}, our scheme with continuous search space tends to be more efficient.

\subsection{Auto machine learning and neural architecture search}

To design a good neural network model for a specified task seems to be always challenging. Recently, Auto Machine Learning (AutoML) aims for saving time and energy for researchers especially for ML beginners. Many companies such as Google Cloud, Amazon AWS also provide capable AutoML services on their cloud products, greatly boosting the democratic process of the AI. 

In the past few years, a new trend of AutoML, neural architecture search (NAS), has also been witnessed to free the researchers from the tedious process of neural architecture design and parameter fine tuning. The early attempts ~\cite {DBLP:journals/corr/abs-1802-03268,DBLP:journals/corr/ZophL16,DBLP:journals/corr/ZophVSL17}
mainly were concentrated on RL to train a controller that can pick out a good sequence of symbols describing the architecture of a neural network. The optimal architecture is a gradual interaction result of action and  environment, which is confined by the search space. These methods achieved great success in image classification task and immediately draw a lot of attention.  Following these seminal works, some alternative approaches ~\cite{DBLP:journals/corr/abs-1802-01548,DBLP:journals/corr/XieY17} that are based on evolutionary algorithms find their applications in this new topic. Evolutionary methods  utilize genetic operations such as selection, mutation and crossover, to filter out weak individual network and yield more competitive ones in the whole process. In addition, it is also capable of handing multi-objective function optimization simultaneously because of the capable method of NSGII~\cite{IEEE:journals/TEC/Deb02}. Therefore, it is widely used for NAS ~\cite{2019arXiv190911409S,DBLP:journals/corr/abs-1904-00420}. However, the above methods tend to require massive computational overhead to complete its whole search process in the large search space for most networks. To overcome this bottleneck, some new methods ~\cite{DBLP:journals/corr/abs-1806-09055,DBLP:journals/corr/abs-1808-07233} have been proposed to efficiently speed up the search process. The authors in ~\cite{DBLP:journals/corr/abs-1806-09055} apply the coefficients to represent the importance of one operator for building a cell for the whole network architecture, and then they relax these coefficients to continuous space using softmax function. The architecture of a good network can be searched at the same time with the weight training process. Luo \etal ~\cite{DBLP:journals/corr/abs-1808-07233} utilize an encoder to map the network architecture to a continuous space and a predictor to estimate its accuracy. Finally, a decoder is taken to obtain the desired architecture. Our auto pruning strategy is motivated by the above great works but our automatic search process is more efficient due to the specified search space design and no multiple mixture operations as in ~\cite{DBLP:journals/corr/abs-1806-09055}.

\section{Methodology}
In this section, we first introduce the formulation of our AutoPruning problem, and then present a framework to solve this problem in an online manner, followed by the details of the whole training pipeline.

\subsection{Problem formulation}
Choosing the number of channels for each layer is not an easy task, and some early works mainly focus on evaluating the importance of each channel, which are built based on different measurements such as square values of weights and so on. To achieve the goal of automatic pruning for different layer, we first formulate the pruning process as a two objectives' optimization problem of hyper parameters of remaining ratio and the network's weight.

Given a network model $M(w,R)$ that is associated with a weight matrix $w$ and a parameter vector $R$ that is directly related to channels, training set $X_{T}$ and validation set $X_{V}$, the target of auto pruning is to find a good combination of channels in each layer such that the validation accuracy is maximized with respect to $R$ and $w$, meanwhile, the corresponding weight $w$ of the model is derived by minimizing the training loss. As such, for the whole neural network it can be described as an optimization problem that maximizes the accuracy on the validation data $X_{V}$ and minimization of the training loss on the training data  $X_{T}$ with following two objectives~\cite{2017arXiv170506270S}:

\begin{equation}
\label{eq:objective_function}
\begin{split}
& \text{max}_{R,w} Acc_V(R,W^*_R))  \\
& \text{s.t.} \,  W^*_R = \text{argmin}_WLoss_T(R,W))     
\end{split}
\end{equation}
where $W^*_{R}$ is the optimal weights associated with $R$, $Acc_{V}(R,W^*_{R})$ and $Loss_{T}(R,W)$ is the respective accuracy value and training loss function based on the channels and weights with the condition of $R$. $X_{T}$ and $X_{V}$ represents the training and validation dataset respectively. 
This refers to a two objectives' optimization problem, where the channel number based $R$ can be considered as hyper-parameter. The inner level optimization is a typical weight optimization problem in most training process of neural network, while the outer level is the search process for the $R$ under the optimal solution of the inner level.

 To consider both the prediction accuracy and pruning effects, we design the loss function $Loss_T(R,W)$ as a combination of cross entropy and FLOPs as follows:
\begin{equation} \label{eq:loss_function1}
\small
\begin{split}
Loss_{T}(R,W) = Loss_{CE}(R,W) + \alpha\times Loss_{R} 
\end{split}
\end{equation}
where $ Loss_{CE}(R,W)$  is the typical cross entropy function that describes the accuracy of a model and $ Loss_{R}$ is a FLOPs related cost. Adding this FLOPs based regularization item to the loss function can efficiently adjust the model's performance, and  $\alpha$ is a regularization coefficient that controls the trade-off of accuracy and compression. To effectively reflect the FLOPs  change in the optimization process after pruning, 
we assume $ Loss_{R}$ is a continuous formation of the channel related vector $R$ :

\begin{equation} \label{eq:loss_function_general}
\small
\begin{split}
Loss_{R}=Cost(R) 
\end{split}
\end{equation}
$Cost(R)$ is a function that describes the FLOPs change of the whole model in the pruning process. Specifically, we assume $R$ represents the remaining ratio of the channels in each layer. More details of this function will be elaborated later.

\subsection{Online optimization}

\noindent The goal of structural pruning is to derive the optimal number of channels in each layer for a neural network. Intuitively speaking, applying AutoML to solve this problem is to search for the discrete combination of different number of channels in different layers, however, such a search space is really huge even for ResNet18, not to mention the much deeper neural network such as ResNet56 and ResNet101. Fortunately, we have cast the auto pruning process into a two objectives' optimization problem, which first free us from the dilemma of discrete search space. Solving it with an exact solution requires calculating high order derivatives and it is very challenging to achieve. To overcome this issue, we make use of an approximated iterative solution and then we can apply the optimization strategy in weight and channel space to simultaneously update $W_{R}$ based on the training losses from $D_{T}$ and renew $R$ based on the validation losses from $D_{V}$ ~\cite{2017arXiv170301785F}.

Assuming $T = 1,2, \dots, T_{max}$ represents the update step of the outer loop and $i = 1, 2, \dots, I $ denotes the update steps of the inner loop. In each step, the network model is trained with a batch size of $B$ images. For the inner loop, we can have:

\begin{equation} \label{eq:update_weight}
W^{(i)}_{T}=W^{(i-1)}_{T-1}-\lambda _{w}\nabla_W\ L_{B}(C,W)
\end{equation}
where $\lambda _{w}$ is the learning rate for the network's weights,  $L_{B}(C, W)$ is the batched loss, and $\nabla_W$ is the gradient with respect to  $w$.
While for the outer loop, we renew the hyper parameters of $R$ per step as follows:
\begin{equation} \label{eq:update_channel}
R_{T} = R_{T-1}-\lambda _{R}\nabla_R\ L_{B}(C,W)
\end{equation}
where $\lambda _{R}$ is the learning rate for the hyper parameters of the remaining ratios in each layer, and $\nabla_R$ is the gradient with respect to $R$. By solving this two objectives' optimization problem using this alternating approximation approach, the pruning channel ratios for each layer can be efficiently searched by gradient based optimization approach. Compared to the discrete search space of the channel numbers in each layer, this search method is more efficient because the design of continuous remaining ratio and the gradient based optimization approach.

\subsection{Dynamic masking in search process}

In the above iteration process of the remaining ratio $R$ for each layer, it only outputs the ratio of channels without the index of each channel for each layer. To solve this problem and obtain the concrete index of each channel, we propose a new dynamic masking process based on convolution weight ranking mechanism to efficiently prune the channels in each layer. Assuming the weights of a convolution for channel $j$ in layer $i$ is $W^{i}_{cov,j}$, we utilize sum of the absolute value of these weights to estimate the importance of one channel, which is also widely used in previous weight pruning or channel pruning related approaches. Then we can rank each channel based on this summary, and the larger it is, the more important this channel is. It should also be pointed out that we do NOT implement this ranking in each iteration step of the whole optimization process. This is because ranking after some iterations tends to produce a more stable and more reliable of evaluation of the importance of each channel. In addition, evaluating the importance of each channel at some intervals will reduce the computation burden of the whole algorithm.

\begin{figure}[t]
\centering
\includegraphics[height=8.68cm]{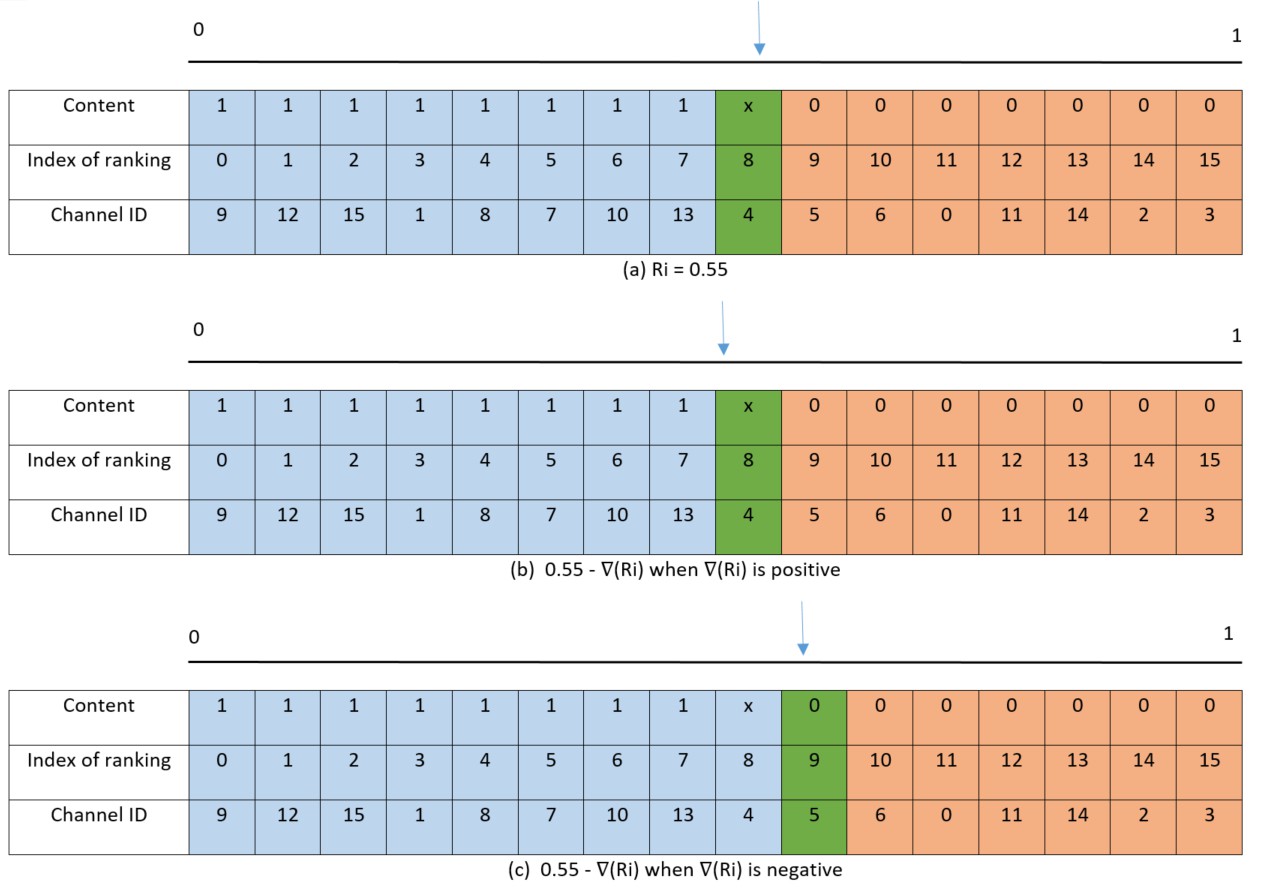}
\caption{Dynamic masking  process. For each subfigure, the first row in the table shows the content of the mask, the second row is its corresponding index of the weight ranking and the third row is its original channel ID in the convolution. The black line above each table means a line that represents the continuous whole remaining ratio range( 0 $\sim$ 1) , and the light blue arrow on each black scale means the current remaining ratio of this layer, i.e., $R_{i}$.  (a) current mask when $R_{i} = 0.55$, (b) possible next mask when the $R_{i}$ moves backward, corresponding to channel shrinkage case (c) possible next mask when the $R_{i}$ moves forward, corresponding to channel enlargement case.(Best viewed in color version)}
\label{fig:mask}
\end{figure}

Based on this ranking, we can further define a mask. For example, if a convolution layer has 16 channels, then this mask is initialized with all 1 at the beginning. In the iteration process, assuming  $R_{i}* C_{i} - floor(R_{i}* C_{i}) = x $ represents a special number that is used for masking purpose, and $C_{i}$ represents the number of channels in this layer and $floor$ is the typical floor function in mathematics, we can define the following mask construction rule:

\begin{equation} \label{eq:mask}
M_{i}(k) = 1 - ReLU(1- ReLU(1+R_{i}*C_{i}-I_{i}(k)))
\end{equation}
where $M_{i}(k)$ represents the mask value of the $k^{th}$ channel in this convolution layer $i$ and $k \in [0, C_{i} - 1]$, and  $ReLU$ is the widely used ReLU function for neural network,  and $I_{i}(k)$ is the importance ranking number of the $k^{th}$ channel in layer $i$ of the DNN.
The above process can be further simplified into the following form:
\begin{equation} \label{eq:mask_rule}
 M_{i}(k) =\left\{
\begin{array}{rcl}
1  &      & {I_{i}(k) \leq floor(R_{i}*C_{i})} \\
x  &      & {I_{i}(k) \in (floor(R_{i}*C_{i}), ceil(R_{i}*C_{i}))} \\
0  &      & {I_{i}(k) \geq ceil(R_{i}*C_{i})}

\end{array} \right. 
\end{equation}
where  $ceil$ is the typical ceiling function used in mathematics.
As shown in Fig.\ref{fig:mask}, the mask change process can also be illustrated with a specific case. Assuming the number of channels is 16 and $R_{i} = 0.55$ for one layer in some epoch of the whole searching process, then the corresponding mask will be $[1,1,1,1,1,1,1,1,x,0,0,0, 0,0,0,0]$, which is shown in Fig.\ref{fig:mask}(a). In the next iteration, the possible mask will be Fig.\ref{fig:mask}(b) or Fig.\ref{fig:mask}(c).
With this mask, we need to map this ranking back to the original channel ID, $M^{'}_{i}(k)$. Then we can further obtain the pruned channels as follows:
\begin{equation} \label{eq:pruning_channel}
F^{i}_{remaining} = F^{i}_{original} \cdot M^{'}_{i}(k)
\end{equation}
where $F^{i}_{original}$ and $F^{i}_{remaining}$ represents the original convolution channels in layer $i$ and the remaining channels after pruning with the re-indexed mask $M^{'}_{i}(k)$

In the iteration process of each channel, we also apply extra rules to ensure the stability of this process, for those channels that have been removed in previous step, we will keep them instead of discarding them immediately. In some cases $R_{i}$ may increase in the whole search process, and the previous channel may still need to provide supplementary channel features in this process. As such, keeping them can enable a more stable iteration process. It can also save the computational time since we don't need to retrain the model to obtain the needed features from those removed channels.


\subsection{Training pipeline}

The whole dynamic channel search for auto pruning based on the collaborative optimization of hyper parameters of the remaining ratios $R$ and the network's weight $W$ is described in Algorithm \ref{alg}. With the training and validation set, initialized remaining ratio value and a pretrained model as inputs, our auto pruning algorithm first calculates the loss based on the training data for forward process, and then it updates the original weight by gradient descent algorithm. For the subsequent validation, it also obtains the validation error, then our algorithm utilizes this error to update the remaining ratios for each convolution layer with gradient descend algorithm. Then, it updates the ranking of the weights at some intervals together with the dynamic mask change.  The outputs of the whole algorithm are the remaining ratio of the channels and its corresponding index for each layer of the whole network model.

\begin{algorithm}[t]
\hrulefill
\caption{Dynamic Channel Search for Auto Pruning}
\label{alg}
\SetKwInOut{Input}{input}\SetKwInOut{Output}{output}
\Input{The training set $X_T$, validation set $X_V$, initialized channels' remaining ratios $R$, pretrained model $M$ with weight $W$. }

    \While{not reaching the target iterations or not converge}{
  Sample data from training data $D_T$\\
  Calculate the loss $L$ on training data with Eq.(\ref{eq:loss_function1})\\
  Update the weight $W$ by gradient descent with Eq.(\ref{eq:update_weight})\\
  Sample data from validation data $D_V$\\
  Calculate the loss $L$ on validation data with Eq.(\ref{eq:loss_function1})\\
  Update the remaining ratio of channels $R$ by Eq.(\ref{eq:update_channel})\\
  Update the ranking of convolution weights for each channel at some iteration intervals followed by dynamic mask update\\
  
 }
    \Output{Remaining ratios of channels and their corresponding index for each layer}
\hrulefill
\end{algorithm}

 As for the loss function in Eq.(\ref{eq:loss_function_general}), we propose the following new function for the sake of convenience of implementation: 

\begin{equation} \label{eq:loss_function10}
\small
\begin{split}
Cost(R) =(\frac{ \sum_{i=1}^{N}P_{i}\times R_{i}}{\sum_{i=1}^{N}P_{i}})^{\beta} 
\end{split}
\end{equation}
where $\beta$ is a positive coefficient, and $P_{i}$ is defined as the FLOPs of the $i^{th}$ layer. 
The FLOPs of a layer i.e., $P_{i}$ , may be regarded as a constant for lay $i$. Since $R_{i}$ is a continuous remaining ratio,  the above equation may be considered differentiable with respect to $R_{i}$.

\section{Experiments}

In this section, we introduce the details of our experiments first, then extensive tests on CIFAR10 together with their analysis are presented to show the validity of the proposed auto pruning scheme.

\subsection{Implementation details}

 In our experiments, we utilize ResNet18, ResNet20, ResNet34, ResNet50, ResNet56 ~\cite{He2015DeepRL} or MobileNetV2 ~\cite{DBLP:journals/corr/abs-1801-04381} as the baseline models for image classification task to test the performance of our auto pruning scheme. 
 For the initialization of the remaining ratios in each layer, all of them are set with 1 for the convenience of iteration. When the optimal pruned network are obtained by the proposed scheme, the model may be further fine tuned to get its final accuracy, which is a practical tradition in the field of model pruning.

As for the parameter $\alpha$ in Eq.(\ref{eq:loss_function1}) and $\beta$ in Eq.(\ref{eq:loss_function10}), a rather suitable set of them is set with (0.5, 0.3) after trials. Concerning the learning rate schedule of weight and remaining ratios of channels, we apply cosine annealing method to adjust them. For cosine annealing in the auto pruning process of the proposed scheme, we also set the $T_{max} = epochs/5 $ for all of our tests, where $epoch$ is the number of our training epoch times. The reason why we set this hyper-parameter in such a manner is because we find that cycling the learning rate this way may yield better accuracy and pruning ratio. In addition, the good results tend to be found by our AutoML approach earlier compared to other choices, saving us a lot of time in tests. The minimum learning rate for them is 0.001 and 0.0001 respectively. The weight ranking iteration interval is set to 800 after trials. We implemented all our experiments with Pytorch on NVidia-V100 with 8 GPUs.

In all the tests, we record the Top-1 prediction accuracy of the pruned model together with the FLOPs pruning ratio (FPR) to show the performances of pruning. A more accurate neural network model with a larger FPR is always desired.

\subsection{CIFAR10 results}
The CIFAR10 dataset is a typical benchmark dataset that consists of 60000 colour images in 10 classes, with 6000 images per class. It is widely applied for image classification test. In this subsection, we compare the proposed algorithm to the method of pruning filter via geometric median (FPGM) in \cite{DBLP:journals/corr/abs-1811-00250}, the method of channel pruning (CP) in \cite{DBLP:journals/corr/HeZS17}, soft filtering pruning (SFP) \cite{DBLP:journals/corr/abs-1808-06866} and AMC \cite{DBLP:journals/corr/abs-1802-03494}. It should be noted that we mainly compare the proposed method to filter pruning based approach since they belong to the same category related methods. In addition, we also do NOT compare knowledge distillation related methods such as ~\cite{pmlr-v97-koratana19a} or pruning followed by knowledge distillation approach in this work. In this part, we take ResNet20, ResNet32 and ResNet56 as examples to show the comparison for the effects of model pruning. 

The auto pruning result of our algorithm and the other related methods are demonstrated in Table\ref{table:CIFAR10}. For ResNet20, our auto pruning algorithm yields much better classification accuracy of 92.06 \% as well as a much higher FPR of 48.35\% compared to SFP and FPGM. On ResNet32, a similar conclusion can also be drawn since the proposed algorithm achieves an impressive accuracy of 92.60\% and a FPR of 45.05\%, outperforming SFP and FPGM in both accuracy and pruning ratio. For ResNet56, our method achieves a better accuracy of 93.51\% over the previous methods such as CP, AMC, SFP and FPGM, together with a comparable FPR of 50.00\% over these three methods.    

Compared to CP that considers the redundancy among inter feature maps, the reason of our superior performance might due to the fact that we directly take into account the accuracy and the FLOPs compression at the same time, leading to a better performance for DNN pruning. SFP method also allows the pruned filters to be updated and yields a large model capacity just as our proposed scheme in this work. However, its dynamic pruning process is mainly based on the gradual measurement of $l_{2}$ norm. The explicit alternating optimization of weight and pruning ratio suggested in this work provides a good    collaborative  optimization effects, leading to better dynamic control of the pruning for DNN compression. FPGM also considers pruning based on redundancy and applies an interesting idea of geometric median to remove the those filters with redundant information. However, such a redundancy based rule might yield a model that has not so high capacity compared to our learning based approach. In addition, the dynamic channel pruning process advanced in this paper also ensures that the pruned filters which still contain some useful features may have a possibility to be kept in the subsequent pruning procedures.
As for the seminal work of AMC, the reason of the better performance for our method might be due to a continuous search space that enables a more complete coverage of model capacity space compared to the discrete one in AMC.

\setlength{\tabcolsep}{4pt}
\begin{table}
\begin{center}
\caption{Comparison of different pruning algorithms on CIFAR10}
\label{table:CIFAR10}
\begin{tabular}{lllll}
\hline\noalign{\smallskip}
Model & Method & Top 1 accuracy & Accuracy drop & FPR \\
\noalign{\smallskip}
\hline
\noalign{\smallskip}
ResNet20  & SFP & { 90.83\% } & 1.37\% & 42.20\%\\
ResNet20  & FPGM & { 91.09\% } & 1.11\% & 42.20\%\\
ResNet20  & Ours& { \bf 92.06\% } & \bf0.64\% & \bf48.35\%\\
\hline
ResNet32  & SFP& { 92.08\% } & 0.59\% & 41.50\%\\
ResNet32  & FPGM& { 92.31\% } & 0.32\% & 41.50\%\\
ResNet32  & Ours& { \bf92.60\% } & \bf0.10\% & \bf45.05\%\\
\hline
ResNet56  & CP & { 91.80\% } & 1.00\% & 50.00\%\\
ResNet56  & AMC & { 91.90\% } & 0.90\% & 50.00\%\\
ResNet56  & SFP & { 93.13\% } & 0.24\% & 52.60\%\\
ResNet56  & FPGM & { 93.49\% } & 0.10\% & \bf52.60\%\\
ResNet56  & Ours& { \bf93.51\% } & \bf0.09\% & 50.00\%\\
\hline
\end{tabular}
\end{center}
\end{table}
\setlength{\tabcolsep}{1.4pt}

\section{Conclusions}

To compress the model of DNN which are large and computational intensive, we have proposed an efficient auto pruning algorithm that is based on search of the optimal channel numbers in each layer and its corresponding weights. Specifically, a dynamic masking process that is based on the ranking of weights for every channel in each layer is utilized to describe the channel evolving pruning process of the network. A two objectives' optimization is carried out alternatively to obtain the solution of it and its corresponding weights.  This AutoML process can efficiently obtain the channel numbers due to the dynamic search space. In addition, we also put forward a new cost function that can control the balance of accuracy and pruning ratio in the whole procedure. Extensive experimental results on CIFAR10, demonstrate the impressive performance of this new scheme. Future work may lie in further refinement of the channel selection based on some properties such as the loss change for the whole network.

\bibliographystyle{unsrt}  



\end{document}